\documentclass[runningheads]{llncs}

\usepackage{accv}

\usepackage{accvabbrv}

\usepackage{graphicx}
\usepackage{booktabs}
\usepackage{algorithm}
\usepackage{algpseudocode}

\usepackage[accsupp]{axessibility}

\usepackage{hyperref}

\usepackage{orcidlink}

\begin{document}

\title{Learned Suppression for 3D Keypoint Detection with a Graph-Transformer Backbone} 

\titlerunning{Learned Suppression for 3D Keypoint Detection}

\author{Batuhan Arda Bekar\orcidlink{0009-0002-7998-3793} \and
Can Sarı\orcidlink{0009-0006-7826-450X} \and
Hüseyin Can Gülkan\orcidlink{0009-0000-6997-2770} \and
Barış~Özcan\orcidlink{0000-0001-8598-1239}}

\authorrunning{B.~A.~Bekar et al.}

\institute{Department of Computer Engineering, Bahçeşehir University, Istanbul, Turkiye\\
\email{\{batuhanarda.bekar, can.sari, huseyin.gulkan\}@bahcesehir.edu.tr}
\email{baris.ozcan@bau.edu.tr}}

\maketitle

\begin{abstract}

Detecting 3D keypoints is a long-standing challenge in computer vision. Most detectors end with a heuristic post-processing step that is not learned. We propose a 3D keypoint detector that improves on this step with a learned suppression module, paired with a Point Transformer backbone that we extend with a directional graph neural network. The module is a graph network over candidates that learns which to keep, which to suppress, and how to relocate the remaining ones. Paired with three backbones, it improves over DBSCAN and greedy non-maximum suppression, and because it operates on candidate features rather than raw geometry, the same formulation applies to both structural and semantic keypoints. Our model surpasses the per-category trained KeypointDETR on 12 of 16 KeypointNet categories, attains the best Corner F1 on the Building3D Entry-Level benchmark, and remains competitive with BWFormer on the larger Tallinn split. GitHub implementation: https://github.com/cansdev/learned-suppression-3d.
\end{abstract}

\section{Introduction}
\label{sec:intro}

\begin{figure}[tb]
  \centering
  \includegraphics[width=0.7\linewidth]{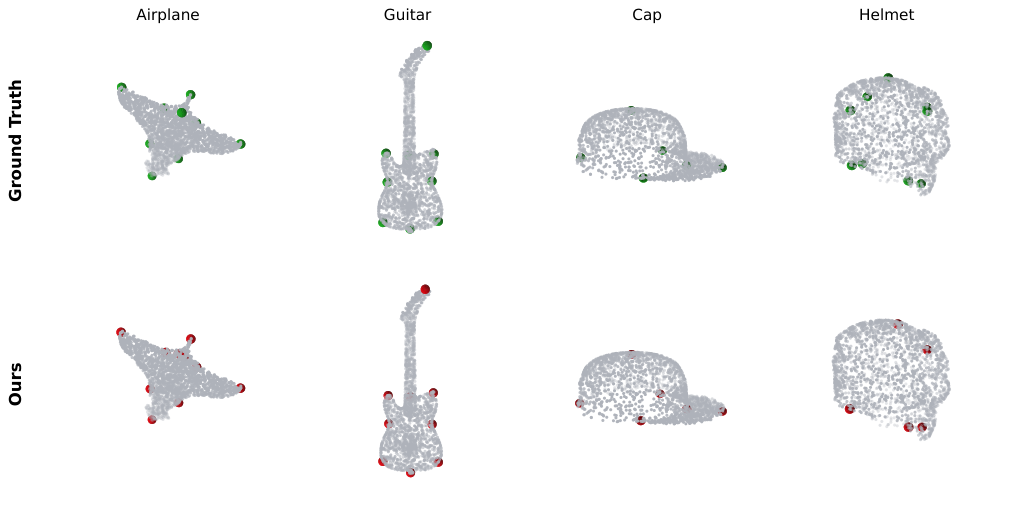}
  \caption{Qualitative results on KeypointNet (Airplane, Guitar, Cap, Helmet): predictions (red), ground truth (green).}
  \label{fig:qual_keypointnet}
\end{figure}

Detecting 3D keypoints is a long-standing challenge in vision and graphics, underpinning object detection, shape matching, retrieval, and registration~\cite{jin2024keypointdetr,you2020keypointnet}. Unlike 2D images, point clouds are unordered, irregularly sampled, and sparse, so each keypoint must be recovered from a small subset of the points in a cloud.

Most learned 3D keypoint detectors predict a dense saliency or vote field over the points; a separate post-processing step, typically non-maximum suppression (NMS) on heatmaps or DBSCAN clustering on votes~\cite{neubeck2006efficient, 3001460.3001507}, then refines this field into a small set of discrete keypoints. This step is external to training: the suppression decision receives no supervision, and the final keypoints are shaped by the heuristic as well as by the features that precede it.

Our approach learns this decision, formulating keypoint extraction as candidate elimination on a graph. Hosang~\etal~\cite{hosang2017learning} showed that NMS can be learned, replacing the fixed deletion rule with a network in which each detection exchanges messages with its neighbors over several rounds and is rescored jointly with them. That network only rescores: detections keep their boxes, and a threshold on the new scores produces the final set. In 3D, rescoring alone sets a limit on localization accuracy: a keypoint rarely coincides with a candidate, and the correct location is often a nearby position that no candidate occupies. Our learned suppression module therefore both suppresses and relocates: a graph over candidate keypoints decides, over several rounds of message passing, which candidates to keep, and moves the remaining ones between rounds so that later rounds operate on more accurate positions.

The learned suppression module operates only on the candidates it receives, so their quality bounds the accuracy it can reach. We therefore build the backbone on a hierarchical Point Transformer encoder~\cite{zhao2021point} and extend it in two places: inside each attention block, a spectral positional encoding conditions attention on the local geometry, and after the decoder, a directional graph neural network (GNN) passes signed messages over the feature contrasts between neighboring points to reinforce geometrically distinctive points such as corners.

\begin{figure}[tb]
  \centering
  \includegraphics[width=0.7\linewidth]{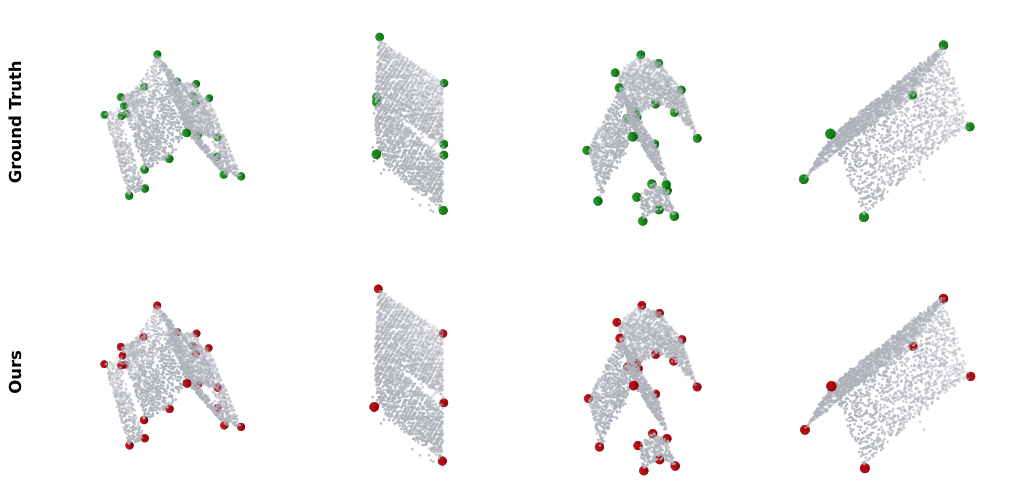}
  \caption{Qualitative results on Building3D (Tallinn): predictions (red), ground truth (green).}
  \label{fig:qual_building3d}
\end{figure}

We evaluate on two datasets with different types of 3D keypoints: the geometric roof corners of Building3D, whose point clouds come from airborne LiDAR~\cite{wang2023building3d}, and the human-annotated semantic landmarks of KeypointNet, which span 16 object categories~\cite{you2020keypointnet}. The leading method in each domain builds in domain-specific assumptions, such as parametric roof priors and bird's-eye-view 2D projections for buildings~\cite{huang2024pbwr,liu2025bwformer}, or a fixed per-category keypoint set for objects~\cite{jin2024keypointdetr}. Our module operates on candidate features rather than raw geometry, so a single formulation with no domain-specific components applies to both. Our pipeline surpasses the per-category KeypointDETR~\cite{jin2024keypointdetr} on 12 of the 16 KeypointNet categories with a single trained model, and attains the best Corner Recall and F1 on Building3D Entry-Level while remaining competitive with BWFormer~\cite{liu2025bwformer} on the full Tallinn split. \Cref{fig:qual_keypointnet,fig:qual_building3d} show example detections on both benchmarks.

Our contributions are as follows:
\begin{itemize}
  \item A learned, graph-based suppression module that outperforms heuristic post-processing such as DBSCAN or greedy NMS for 3D keypoint detection.
  \item A graph-transformer backbone that extends a Point Transformer encoder with a spectral positional encoding and a directional GNN, which, when paired with the learned suppression module, achieves results that are competitive or better than state-of-the-art detectors on both structural and semantic keypoints.
\end{itemize}

\section{Related Work}
\label{sec:related}

\subsection{Point Cloud Processing and Graph Neural Networks}

Learning directly on point sets started with PointNet~\cite{qi2017pointnet}, which applies shared per-point multi-layer perceptrons (MLPs) and permutation-invariant pooling. To capture global context, PointNet++~\cite{qi2017pointnet++} added a hierarchical set abstraction (SA) encoder that groups and aggregates local neighborhoods at progressively coarser scales, which is the encoder structure we build on. Point Transformer~\cite{zhao2021point} then replaced pooling with local vector self-attention, which our encoder adopts, and Dynamic Graph CNN~\cite{wang2019dynamic} builds edge features from the differences between a point and its neighbors, a contrast formulation that our graph layer extends with explicit edge direction and signed messages. Other point and graph networks (KPConv~\cite{thomas2019kpconv}, PAConv~\cite{xu2103position}, RandLA-Net~\cite{hu2020randla}, Stratified Transformer~\cite{lai2022stratified}, FG-Net~\cite{liu2022fg}, PRA-Net~\cite{cheng2021net}) explore alternative convolution, sampling, and attention schemes; we use them as baselines for comparison. A separate line of work pretrains point cloud representations without labels~\cite{yu2022point,pang2023masked,zhang2022point,zhou20243d}. These representations transfer well to classification, but they fall behind supervised training when the task requires precise localization, as our Building3D results confirm.

\subsection{3D Keypoint Detection}

Classic 2D detectors such as SIFT~\cite{lowe2004distinctive}, SURF~\cite{bay2006surf}, and ORB~\cite{rublee2011orb} select interest points by handcrafted local criteria. Harris 3D~\cite{sipiran2011harris} and learned descriptor-specific detectors~\cite{salti2015learning} carry this notion of geometric saliency to meshes and point clouds, spreading points across the surface without semantic bias.

KeypointNet~\cite{you2020keypointnet} introduced a large-scale benchmark of human-annotated semantic keypoints over 16 ShapeNet categories, and showed that learned detectors outperformed geometric ones in semantic keypoint detection. KeypointDETR~\cite{jin2024keypointdetr} treats keypoint detection as set prediction with a transformer decoder and predicts per-query heatmaps to avoid explicit clustering, but requires a fixed per-category keypoint set. Others forgo landmark supervision: Skeleton Merger~\cite{shi2021skeleton} merges skeletal hypotheses, UKPGAN~\cite{you2022ukpgan} distills saliency through reconstruction and adversarial losses, and Wei~\etal~\cite{wei2021multi} jointly predict saliency and cross-shape correspondence. Many of these methods end with a heuristic post-processing step such as greedy NMS.

Hosang~\etal~\cite{hosang2017learning} showed that this step can be learned by jointly rescoring detections. We bring learned suppression to 3D keypoints, extending it to move and refine the remaining keypoint locations instead of only rescoring them in place.

\subsection{Structural Building Analysis from LiDAR}

Learned detection of building structure from airborne LiDAR has progressed from synthetic to real, urban-scale data. PC2WF~\cite{liu2021pc2wf} detects structural corner points directly from raw point clouds, but only on synthetic, rectilinear shapes. Point2Roof~\cite{li2022point2roof} was the first to learn roof modeling end to end, regressing roof corner points and consolidating them with DBSCAN clustering. Building3D~\cite{wang2023building3d} introduced an urban-scale airborne LiDAR benchmark with annotated roof corners, and we adopt its corner-detection task and metrics. More recent methods raise precision by encoding building-specific priors: PBWR~\cite{huang2024pbwr} fits parametric roof typologies, and BWFormer~\cite{liu2025bwformer} detects corners in a 2D top-down projection and lifts them to 3D with a transformer. These priors are specific to roof shapes and do not apply to object-level keypoints.  BuildingWorld~\cite{huang2026buildingworld} argues that the limited architectural and geographic diversity of current datasets constrains the generalization of learned building models, which motivates our approach.

\section{Method}
\label{sec:method}

\begin{figure}[tb]
  \centering
  \includegraphics[width=\textwidth]{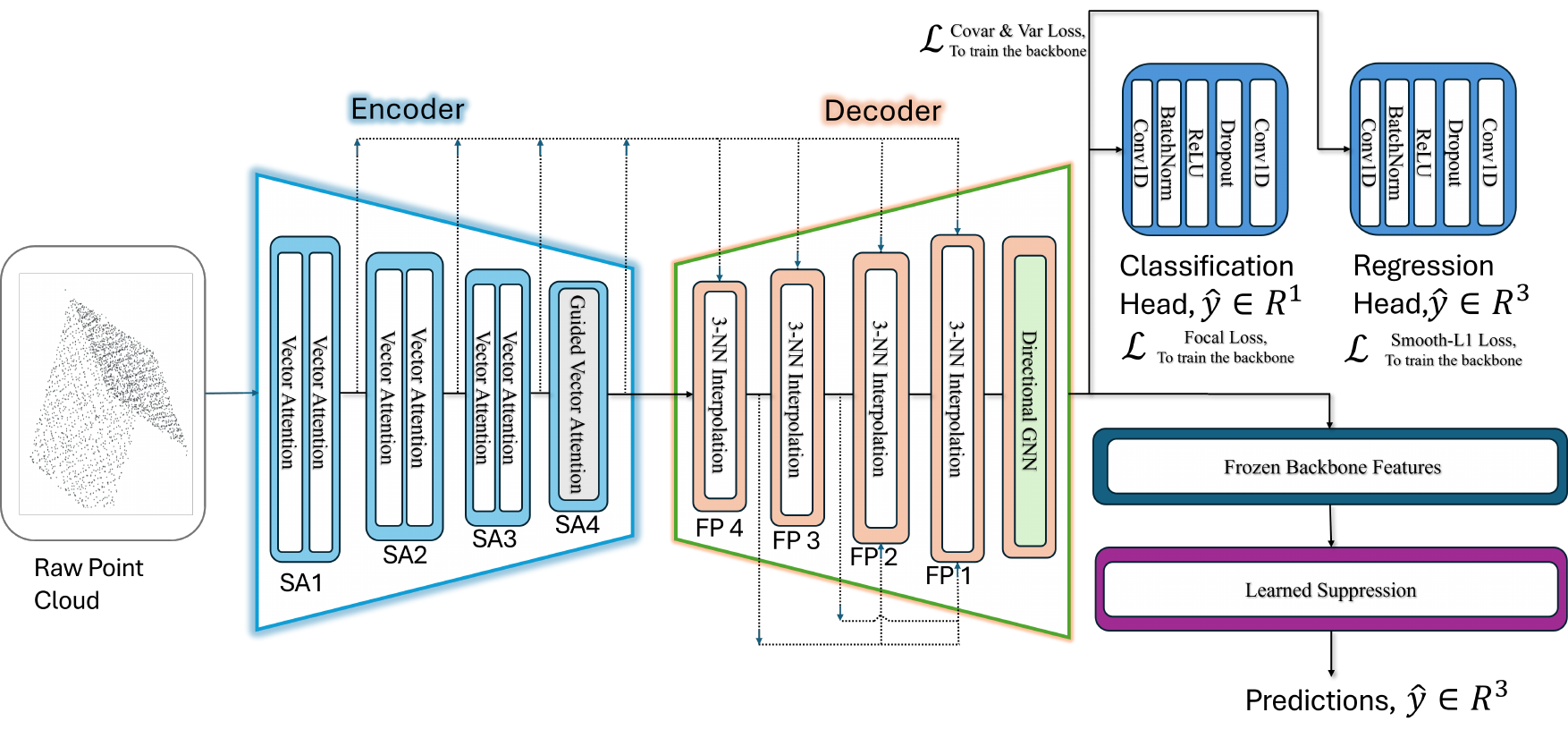}
  \caption{Model architecture. A hierarchical Point Transformer encoder (SA1--SA4; SA4 injects a ground truth log prior at training time only) feeds a UNet3+ decoder (FP1--FP4) with dense cross-scale skips; a directional GNN refines the full-resolution features before the classification and offset regression heads, whose candidates our learned suppression module consolidates into the final keypoint predictions $\hat{y}\in\mathbb{R}^3$ at inference (\cref{sec:postproc}).}
  \label{fig:architecture}
\end{figure}

Our pipeline has two stages (\cref{fig:architecture}). A hybrid backbone, a hierarchical Point Transformer encoder (\cref{sec:encoder}) and a UNet3+ decoder with a directional GNN (\cref{sec:decoder}), extracts per-point features trained to predict dense classification scores and spatial offsets (\cref{sec:loss}). High-confidence predictions then vote to form a candidate set, which our learned suppression module (\cref{sec:postproc}) iteratively suppresses and relocates into the final keypoints.

\begin{figure}[tb]
  \centering
  \begin{subfigure}[t]{0.48\textwidth}
    \centering
    \includegraphics[width=\textwidth]{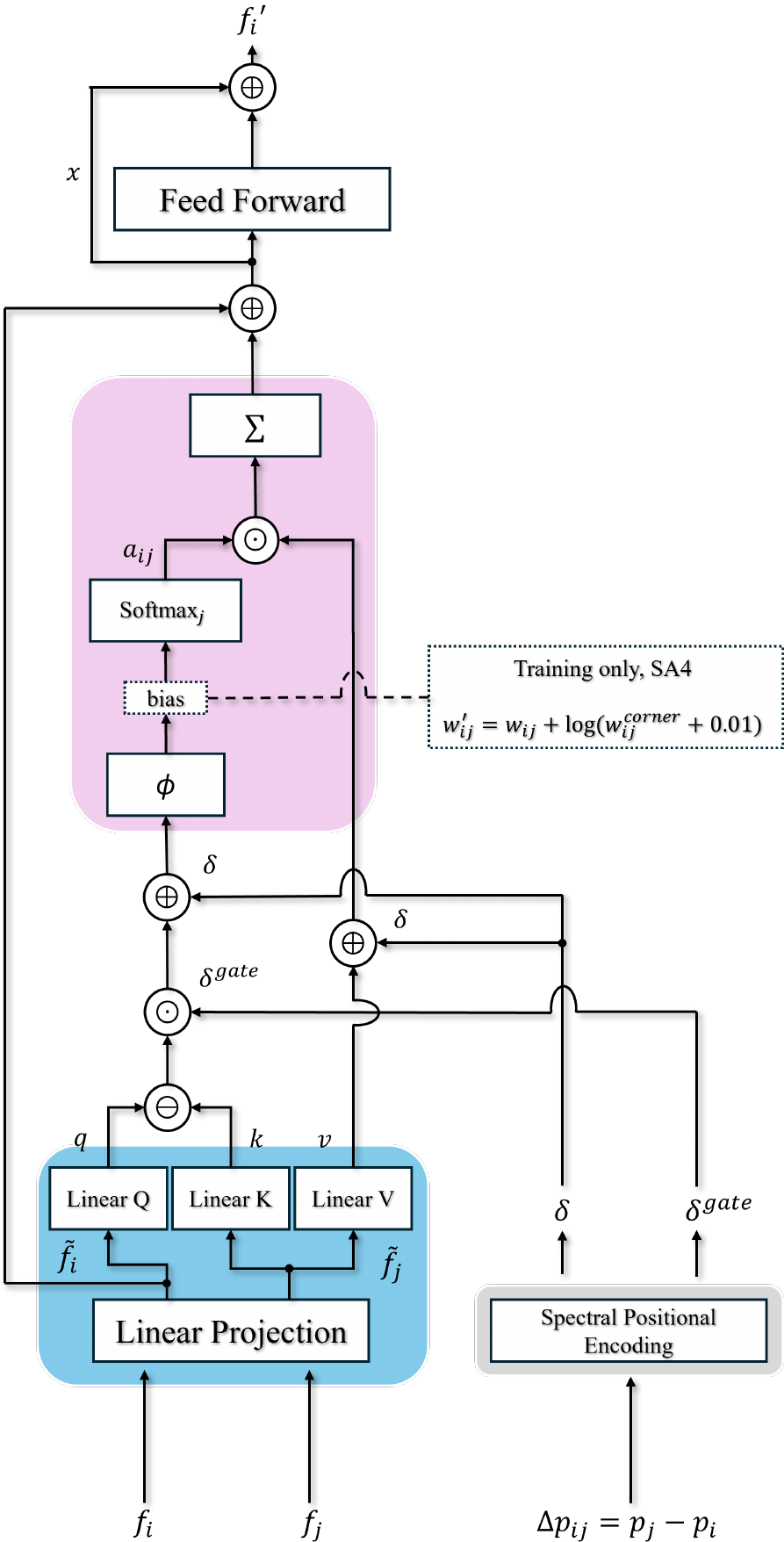}
    \caption{}
    \label{fig:pt_block_sub}
  \end{subfigure}
  \hfill
  \begin{subfigure}[t]{0.48\textwidth}
    \centering
    \includegraphics[width=\textwidth]{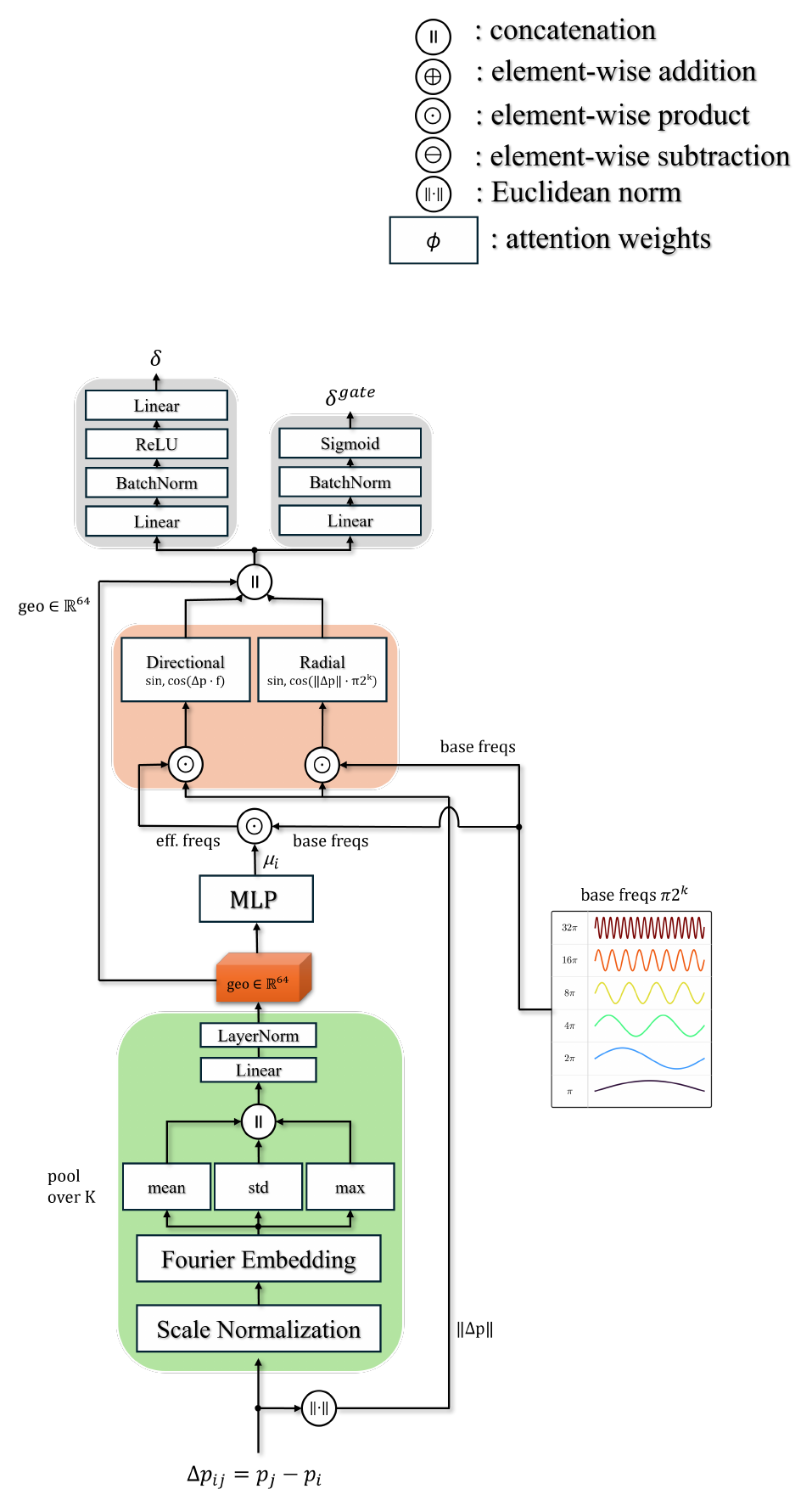}
    \caption{}
    \label{fig:pos_encoder_sub}
  \end{subfigure}
  \caption{\textbf{(a)}~Point Transformer block (\cref{sec:encoder}); the dashed callout is the SA4 training-only log prior of \cref{eq:sga}. \textbf{(b)}~Spectral positional encoding (\cref{eq:pe}).}
  \label{fig:pt_block}
\end{figure}

\subsection{Hierarchical Encoder: Point Transformer Blocks}
\label{sec:encoder}

The encoder is a stack of four set abstraction stages, SA1 to SA4, each of which downsamples the cloud, grows the feature dimension, and applies two vector attention Point Transformer blocks in which a center point aggregates its $k$ nearest neighbors (\cref{fig:pt_block}). 

Inside a block, a shared linear map with no bias first projects the center feature $f_i$ and each neighbor feature $f_j$; three further maps with no bias then produce a query $q_i$, key $k_j$, and value $v_j \in \mathbb{R}^{C}$, where $C$ is the feature dimension at that stage (\cref{fig:pt_block}).

A vector attention block relates the center and its neighbor in feature space only, with no information about where the neighbor lies relative to the center. We supply this with a relative-position term built from the offset $\Delta p_{ij} = p_j - p_i$, where $p_i$ and $p_j$ are the coordinates of the center and the neighbor. Since coordinate-based MLPs are biased toward low-frequency functions and represent fine geometric detail poorly~\cite{tancik2020fourier}, we lift the offset into a sinusoidal feature space with the $[\sin,\cos]$ Fourier-feature map of~\cite{tancik2020fourier}, using the octave band schedule $\omega^{\mathrm{base}} = (\pi, 2\pi, \dots, 2^{L-1}\pi)$ of NeRF-style positional encoding~\cite{mildenhall2021nerf}, where $L$ is the number of frequency bands. A directional encoding $\gamma^{\mathrm{dir}}_{ij}$ applies this map channel-wise to $\Delta p_{ij}$, and a radial encoding $\gamma^{\mathrm{rad}}_{ij}$ applies it to the length $\lVert\Delta p_{ij}\rVert$.

Fixed bands cannot adjust to the size and shape of each local neighborhood. Following the learnable-frequency idea of~\cite{li2021learnable}, we make the directional bands geometry-adaptive: a $64$-dimensional descriptor $g_i$ of the neighborhood (obtained by Fourier-embedding the scale-normalized neighbor offsets and pooling them by mean, standard deviation, and maximum over the $k$ neighbors, shown as the ``geo'' vector in \cref{fig:pt_block}) is passed through an MLP that predicts per-axis band multipliers $\mu_i = \mathrm{MLP}(g_i)$. These multipliers rescale the octave bands along each spatial axis (with $\odot$ the element-wise product) to give the effective frequencies $\omega_i$ (shown as eff. freqs in \cref{fig:pt_block}), while the radial bands stay fixed. A further projection $\psi$ maps the two encodings and the descriptor to an additive position bias $\delta_{ij}$ and a multiplicative gate $\delta^{gate}_{ij}$,
\begin{equation}
\omega_i = \omega^{\mathrm{base}}\odot \mu_i, \qquad
(\delta_{ij},\, \delta^{gate}_{ij}) = \psi\big([\,\gamma^{\mathrm{dir}}_{ij},\; \gamma^{\mathrm{rad}}_{ij},\; g_i\,]\big),
\label{eq:pe}
\end{equation}
where the directional encoding $\gamma^{\mathrm{dir}}_{ij}$ is evaluated at the effective frequencies $\omega_i$. This geometry-adaptive map is the spectral positional encoding of \cref{fig:pt_block}b.

The gate $\delta^{gate}_{ij}\in(0,1)^{C}$ multiplies the content difference $q_i - k_j$ channel by channel, attenuating the feature contrast between the center and the neighbor, similar to the multiplicative position term in Point Transformer~v2~\cite{wu2022point}. The bias $\delta_{ij}$ is applied as an offset to the gated relation and then added again into the values. A small per-pair MLP $\phi$ (two $1{\times}1$ layers with BatchNorm and ReLU) maps the gated and biased relation to the attention weights. As in the vector attention of Point Transformer~\cite{zhao2021point}, the attention weights modulate the values $v$ per channel rather than through a single shared scalar; to keep this compact, $\phi$ outputs one weight for every group of $g$ channels, where $g{=}8$. A softmax over the neighbors normalizes them into $\alpha_{ij}$, which aggregate the values, each shifted by the position bias $\delta_{ij}$,
\begin{equation}
\begin{aligned}
w_{ij} &= \phi\big(\delta^{gate}_{ij}\odot(q_i - k_j) + \delta_{ij}\big), \qquad
\alpha_{ij} = \mathrm{softmax}_j(w_{ij}), \\
\mathrm{out}_i &= \sum_j \alpha_{ij}\odot(v_j + \delta_{ij}).
\end{aligned}
\label{eq:vector_attn}
\end{equation}
The aggregated output $\mathrm{out}_i$ is added to the center feature through a residual connection and refined by a feed-forward layer with a second residual (\cref{fig:pt_block}).

This Point Transformer block is distinct at SA4 in training time, the final and coarsest Set Abstraction stage (Guided Vector Attention at \cref{fig:architecture}), where a training-time-only prior following the paradigm of learning using privileged information~\cite{vapnik2009new} is used. We bias the attention logits with a log term derived from the soft ground truth keypoint labels,
\begin{equation}
\begin{aligned}
w_{ij}' &= w_{ij} + \log\!\big(w^{corner}_{ij} + 0.01\big), \\
w^{corner}_{ij} &= y_j, \qquad
y_i = \mathrm{clip}\!\big(\exp(-d^{\min}_i / r),\, 0,\, 1\big),
\end{aligned}
\label{eq:sga}
\end{equation}
where $y_i$ is the soft ground truth label of point $i$, set from its distance $d^{\min}_i$ to the nearest annotated keypoint with radius $r$, and $w^{corner}_{ij}$ carries this label down the sampling hierarchy to SA4 and across each center's neighbors. This prior is used during training only: no ground truth label enters the network at inference, where \cref{eq:sga} reduces exactly to the plain vector attention of \cref{eq:vector_attn}.

\subsection{UNet3+ Skip Connections and Directional GNN}
\label{sec:decoder}

The decoder mirrors the encoder with four feature propagation stages, FP4 to FP1, each of which upsamples the features by three-nearest-neighbor interpolation. Following UNet3+~\cite{huang2020unet}, it fuses dense skip connections from every encoder and decoder stage; \cref{fig:architecture} shows the connectivity.

After FP1, a directional GNN refines the per-point features over a $k$-nearest-neighbor ($k$-NN) graph. On Building3D the graph uses Euclidean neighborhoods; on KeypointNet, where landmarks lie along the object surface, the neighborhood and the edge length $\mathrm{dist}_{ij}$ are instead measured as geodesic distances, which strengthens the contrast features (\cref{tab:ablations}). We follow the edge-feature formulation of Dynamic Graph CNN~\cite{wang2019dynamic}: each edge $(i,j)$ carries an edge feature $e_{ij}$ that concatenates the feature contrast $f_i - f_j$ between a point and its neighbor with the unit edge direction $\mathrm{dir}_{ij}$ and the edge length $\mathrm{dist}_{ij}$,
\begin{equation}
e_{ij} = [\,f_i - f_j,\; \mathrm{dir}_{ij},\; \mathrm{dist}_{ij}\,], \qquad
\mathrm{dir}_{ij} = \frac{p_j - p_i}{\|p_j - p_i\|}, \qquad
\mathrm{dist}_{ij} = \|p_j - p_i\|.
\label{eq:gnn_edge}
\end{equation}
Two MLPs act on each edge feature: $\mathrm{MLP}_a$ produces an attention logit that is normalized over the neighbors $j$ by a softmax (scaled by $1/\sqrt{C}$, with $C$ the feature dimension at the FP1 stage) to give the weight $a_{ij}$, and $\mathrm{MLP}_m$ produces a signed message $m_{ij}$. We aggregate the messages and add them back through a learnable gated residual,
\begin{equation}
\begin{aligned}
a_{ij} &= \mathrm{softmax}_j\!\Big(\tfrac{\mathrm{MLP}_a(e_{ij})}{\sqrt{C}}\Big), \qquad
m_{ij} = \mathrm{MLP}_m(e_{ij}), \\
f_i' &= f_i + \sigma(\lambda)\, \mathrm{MLP}_o\!\Big(\textstyle\sum_j a_{ij}\, m_{ij}\Big),
\end{aligned}
\label{eq:gnn_msg}
\end{equation}
where $\mathrm{MLP}_o$ is an output projection, $\sigma$ is the logistic sigmoid, and $\lambda$ is a single learnable scalar that gates the update; $\mathrm{MLP}_m$ has no final activation, so a message can either increase or decrease a feature's importance. The messages are signed and carry direction, so that points receive different contrasts depending on their positions with respect to their neighbors. The directional GNN outputs refined per-point features, which are then fed as input to the learned suppression module (\cref{sec:postproc}).

\subsection{Learned Suppression}
\label{sec:postproc}

Inspired by three lines of work, our learned suppression module makes the final selection itself learned: a network over a small candidate set, rather than the full cloud, decides which candidates become keypoints and where they lie. Learned NMS replaces a fixed suppression rule with a network that rescores detections from their neighbors~\cite{hosang2017learning}; relation networks remove duplicate detections by modeling pairwise relations between objects~\cite{hu2018relation}; and permutation-invariant set networks process an unordered set with attention and impose no ordering~\cite{lee2019set}. For 3D keypoints, our module is a permutation-invariant graph over candidates that learns pairwise suppression, and it additionally iterates and relocates the remaining candidates between rounds.

\paragraph{Candidate generation.} Every point whose predicted corner probability exceeds a threshold casts a Hough-style vote at its offset-displaced position $p_i + r\,\hat{\Delta}_i$~\cite{qi2019deep}. We denoise the votes with two iterations of mean-shift and initially reduce them to a candidate set by greedy NMS. Each candidate $j$ pools the statistics of its voters together with the refined features from \cref{eq:gnn_msg}, which yields the candidate feature $z_j$ and its position $c_j$.

\paragraph{Learned elimination.} The learned suppression module is a small graph network over the candidate set (\cref{alg:slice}), built from five MLPs. A shared MLP, $\mathrm{MLP}_{\mathrm{emb}}$, first embeds each candidate feature $z_j$ into a hidden state $h_j\in\mathbb{R}^{128}$. The module then runs $T$ rounds of message passing within a single forward pass; a round is one iteration of the module's computation, different from a training epoch. In each round, for every ordered pair of candidates within a radius $\rho$, $\mathrm{MLP}_{\mathrm{rel}}$ maps the two hidden states and their current distance $d_{ij}$ to a suppression relation $s_{ij}\in[0,1]$, the degree to which candidate $i$ suppresses candidate $j$. Each candidate then aggregates the suppression it receives into a message $u_j=\sum_i s_{ij}\,h_i$; $\mathrm{MLP}_{\mathrm{upd}}$ updates its hidden state from $[h_j,u_j]$; and $\mathrm{MLP}_{\mathrm{mov}}$ predicts a displacement that relocates the candidate before the next round. After the final round, a shared $\mathrm{MLP}_{\mathrm{keep}}$ maps each hidden state to a keep score, which is multiplied by $1-\max_i s_{ij}$ (the strongest suppression the candidate received) to give an effective keep probability $\pi_j$. Candidates with $\pi_j$ above a threshold $\tau$ are kept and moved by their respective displacement.

\begin{algorithm}[t]
\caption{Learned suppression module. $\mathrm{MLP}_{\mathrm{emb}}$ and $\mathrm{MLP}_{\mathrm{keep}}$ are shared across rounds; $\mathrm{MLP}_{\mathrm{rel}}$, $\mathrm{MLP}_{\mathrm{upd}}$, and $\mathrm{MLP}_{\mathrm{mov}}$ have independent weights in each round $t$. $\rho$: suppression radius; $T$: message-passing rounds per forward pass; $\tau$: keep threshold.}
\label{alg:slice}
\begin{algorithmic}[1]
\Require candidate features $\{z_j\}$, candidate positions $\{c_j\}$
\State $h_j \gets \mathrm{MLP}_{\mathrm{emb}}(z_j)$ \Comment{shared embedding into $\mathbb{R}^{128}$}
\For{$t = 1,\dots,T$}
  \State $d_{ij} \gets \lVert c_i - c_j \rVert$,\quad $\mathcal{A} \gets \{(i,j): d_{ij} < \rho,\; i \neq j\}$
  \State $s_{ij} \gets \sigma\!\big(\mathrm{MLP}_{\mathrm{rel}}([\,h_i, h_j, d_{ij}\,])\big)$ for $(i,j)\in\mathcal{A}$, else $0$ \Comment{$s_{ij}$: $i$ suppresses $j$}
  \State $u_j \gets \sum_i s_{ij}\, h_i$ \Comment{incoming suppression}
  \State $h_j \gets h_j + \mathrm{MLP}_{\mathrm{upd}}([\,h_j, u_j\,])$
  \State $c_j \gets c_j + \mathrm{MLP}_{\mathrm{mov}}(h_j)$ \Comment{relocate; $c_j$ accumulates the displacement}
\EndFor
\State $\pi_j \gets \sigma\!\big(\mathrm{MLP}_{\mathrm{keep}}(h_j)\big)\cdot\big(1 - \max_i s_{ij}\big)$ \Comment{$s_{ij}$ from the final round}
\State \Return $\{c_j : \pi_j > \tau\}$
\end{algorithmic}
\end{algorithm}

\subsection{Loss Design}
\label{sec:loss}

Each point carries the soft keypoint label $y_i$ from \cref{eq:sga}, and the positive set $P$ collects all points within radius $r$ of an annotated keypoint; for these the regression target is the displacement to the nearest keypoint, scaled by $1/r$. The classification and regression components of our loss are inspired by Point2Roof~\cite{li2022point2roof}.

\paragraph{Classification.} The per-point corner logits $x_i$ are trained with a focal loss~\cite{lin2018focallossdenseobject} to counter class imbalance, with two additions. The focal exponent $\gamma$ is annealed from $2.0$ to a floor of $1.05$ over the first $100$ epochs and then held fixed, so the loss concentrates on rare positives early and relaxes once they are learned. Each point is further reweighted by $1 + \beta\, e^{-d^{\min}_i/r}$, where $d^{\min}_i$ is the distance from point $i$ to its nearest annotated keypoint (as in \cref{eq:sga}), $r$ is the label radius, and $\beta = 2.0$; this weight emphasizes points near a keypoint,
\begin{equation}
\begin{aligned}
L_{cls} &= \mathbb{E}_i\Big[(1 + \beta\, e^{-d^{\min}_i/r})\,(1 - p_{t,i})^{\gamma}\,\mathrm{BCE}(x_i, y_i)\Big], \\
p_{t,i} &= \sigma(x_i)\, y_i + (1-\sigma(x_i))(1 - y_i).
\end{aligned}
\label{eq:cls}
\end{equation}
Here $\mathbb{E}_i$ denotes the mean over all points $i$, $\mathrm{BCE}$ is the binary cross-entropy, $\sigma$ is the logistic sigmoid, and $p_{t,i}$ is the focal loss probability of the true class at point $i$.

\paragraph{Offset regression.} An annotated keypoint rarely coincides with a sampled input point~\cite{wang2023building3d}, so per-point classification alone cannot predict it exactly; for each point in the positive set $P$, the offset regression head predicts displacements to its target keypoint, recovering locations that lie off the input point set. The predicted offsets of the positive set are trained with a Smooth-$L_1$ loss~\cite{girshick2015fast}, summed over the three coordinates and averaged over $|P|$; we denote it $L_{off}$.

\paragraph{Feature regularization.} Because the learned suppression module consumes the post-GNN per-point features, we guard them against representational collapse with the variance and covariance terms of VICReg~\cite{bardes2021vicreg} ($L_{var}$, $L_{cov}$); \cref{tab:ablations} reports their effect.

\paragraph{Total.} The backbone is trained end to end with
\begin{equation}
L = L_{cls} + L_{off} + 0.1\,L_{var} + 0.01\,L_{cov}.
\label{eq:total}
\end{equation}

\paragraph{Two-stage training.} The learned suppression module is trained separately, on top of the frozen backbone, so the candidate features it consumes are exactly those it sees at inference. Candidates are matched to ground truth keypoints by Hungarian assignment~\cite{Kuhn1955Hungarian}; a matched candidate supervises a binary keep loss (cross-entropy with target $1$) and a Smooth-$L_1$ displacement loss toward its matched keypoint, while unmatched candidates are supervised to be suppressed (keep target $0$).

\begin{table}[tb]
  \caption{Comparison with prior methods on Building3D (test split). Best per section in \textbf{bold}; prior results from~\cite{wang2023building3d} (Entry-Level) and~\cite{liu2025bwformer} (Tallinn). $^\dagger$``Supervised (PT)'' uses Point-Transformer~\cite{zhao2021point}; ``Self-supervised''/``Linear self-supervised'' are Point-MAE-based variants from~\cite{wang2023building3d}.}
  \label{tab:building3d_sota}
  \centering
  \resizebox{\linewidth}{!}{%
  \begin{tabular}{@{}lrrrr@{\hspace{1.5em}}lrrrr@{}}
    \toprule
    \multicolumn{5}{c}{\textit{Entry-Level}} & \multicolumn{5}{c}{\textit{Tallinn}} \\
    \cmidrule(r){1-5}\cmidrule(l){6-10}
    Method & ACO$\downarrow$ & CP & CR & CF1 & Method & ACO$\downarrow$ & CP & CR & CF1 \\
    \midrule
    Point-BERT~\cite{yu2022point}         & 0.25           & 88.0 & 69.0 & 77.0 & Point-MAE~\cite{pang2023masked}        & 0.330          & 75.0          & 47.0          & 58.0 \\
    Point-MAE~\cite{pang2023masked}      & 0.27           & 85.0 & 69.0 & 76.0 & Point-M2AE~\cite{zhang2022point}       & 0.320          & 79.0          & 58.0          & 67.0 \\
    Point-M2AE~\cite{zhang2022point}     & 0.26           & 88.0 & 69.0 & 77.0 & Point2Roof~\cite{li2022point2roof}     & 0.390          & 65.0          & 30.0          & 41.0 \\
    3D-OAE~\cite{zhou20243d}             & 0.27           & 86.0 & 68.0 & 76.0 & Linear self-supervised$^\dagger$~\cite{wang2023building3d} & 0.350 & 70.0 & 60.0 & 65.0 \\
    Self-supervised$^\dagger$~\cite{wang2023building3d} & 0.26 & 87.0 & 69.0 & 77.0 & Supervised (PT)$^\dagger$~\cite{wang2023building3d} & 0.290 & 90.0 & 53.0 & 66.0 \\
    PointNet~\cite{qi2017pointnet}        & 0.36           & 71.0 & 50.0 & 59.0 & PC2WF~\cite{liu2021pc2wf}              & 0.520          & 18.0          & 67.0          & 28.0 \\
    PointNet++~\cite{qi2017pointnet++}   & 0.34           & 79.0 & 52.0 & 63.0 & PBWR~\cite{huang2024pbwr}              & 0.222          & \textbf{98.5} & 68.8          & 81.0 \\
    RandLA-Net~\cite{hu2020randla}       & 0.35           & 70.0 & 60.0 & 65.0 & BWFormer~\cite{liu2025bwformer}        & \textbf{0.204} & 94.9          & 82.7 & \textbf{88.4} \\
    DGCNN~\cite{wang2019dynamic}         & 0.32           & 73.0 & 58.0 & 65.0 & Ours                                  & 0.216          & 92.3          & \textbf{83.7}          & 87.8 \\
    PAConv~\cite{xu2103position}         & 0.33           & 75.0 & 57.0 & 65.0 & & & & & \\
    Stratified Transformer~\cite{lai2022stratified} & 0.38 & 72.0 & 51.0 & 62.0 & & & & & \\
    FG-Net~\cite{liu2022fg}              & 0.32           & 77.0 & 64.0 & 70.0 & & & & & \\
    Point2Roof~\cite{li2022point2roof}   & 0.30           & 66.0 & 48.0 & 56.0 & & & & & \\
    Supervised (PT)$^\dagger$~\cite{wang2023building3d} & 0.26 & 89.0 & 66.0 & 76.0 & & & & & \\
    PBWR~\cite{huang2024pbwr}            & \textbf{0.22}           & \textbf{97.0} & 85.0 & 91.0 & & & & & \\
    Ours                                & \textbf{0.22}          & 96.5 & \textbf{90.1} & \textbf{93.2} & & & & & \\
    \bottomrule
  \end{tabular}}
\end{table}

\begin{table}[tb]
  \caption{Component ablations; Building3D Entry-Level numbers are measured on a held-out validation split, a $10\%$ hold-out of the Entry-Level training set. The Op.\ (Operation) column marks whether each row adds a new component, swaps one for another ($x\rightarrow y$), or sets the parameters of a component already in the pipeline to the best configuration on this validation split; the running model carries every change above it. The base of (b) is the full backbone built in (a), the running model after the spectral positional encoding row, with the directional GNN on Euclidean neighborhoods. $^\dagger$\,marks the same running model shared across the two sections. Best per section in \textbf{bold}.}
  \label{tab:ablations}
  \centering
  \resizebox{\linewidth}{!}{%
  \begin{tabular}{@{}llrrrrc@{\hspace{1.5em}}llrr@{}}
    \toprule
    \multicolumn{7}{c}{\textit{(a) Building3D Entry-Level}} & \multicolumn{4}{c}{\textit{(b) KeypointNet (all 16 categories)}} \\
    \cmidrule(r){1-7}\cmidrule(l){8-11}
    Configuration & Op. & ACO$\downarrow$ & CP & CR & CF1 & $\Delta$CF1 & Configuration & Op. & mIoU & CD$\downarrow$ \\
    \midrule
    PointNet++                    & base & 0.311 & 74.7 & 80.9 & 77.66 & --      & Full backbone from (a)$^\dagger$         & base & 77.37 & 0.091 \\
    Contrast GNN                  & add  & 0.308 & 76.5 & 81.9 & 79.10 & +1.44   & Euclidean $\rightarrow$ geodesic graph   & swap & 79.67 & 0.085 \\
    Point Transformer blocks      & add  & 0.300 & 84.2 & 83.9 & 84.03 & +4.93   & DBSCAN $\rightarrow$ NMS                 & swap & 84.25 & 0.080 \\
    SA4 attention prior           & add  & 0.291 & 89.0 & 80.1 & 84.32 & +0.29   & NMS $\rightarrow$ learned supp.\ (relocate \& rescore) & swap & \textbf{92.01} & \textbf{0.055} \\
    DBSCAN grid search             & set  & 0.301 & 86.3 & 83.8 & 85.00 & +0.68   & & & & \\
    Var/Covar regularizer         & add  & 0.298 & 87.8 & 83.8 & 85.72 & +0.72   & & & & \\
    ReLU $\rightarrow$ GELU       & swap & 0.297 & 88.0 & 84.0 & 85.95 & +0.23   & & & & \\
    UNet skip conn. $\rightarrow$ UNet3+ skip conn. & swap & 0.290 & 88.6 & 84.3 & 86.44 & +0.49   & & & & \\
    Edge direction in GNN         & add  & 0.279 & 88.1 & 86.0 & 87.08 & +0.64   & & & & \\
    Spectral positional encoding$^\dagger$ & add  & 0.280 & 90.3 & 86.8 & 88.55 & +1.47   & & & & \\
    DBSCAN $\rightarrow$ learned supp.\ (rescore only) & swap & 0.280 & 95.3 & \textbf{89.9} & 92.50 & +3.95 & & & & \\
    Rescore only $\rightarrow$ rescore \& relocation & swap & \textbf{0.251} & \textbf{96.1} & 89.5 & \textbf{92.70} & +0.20 & & & & \\
    \bottomrule
  \end{tabular}}
\end{table}

\section{Experiments}
\label{sec:experiments}

\begin{table}[tb]
  \caption{Per-category keypoint localization on all 16 KeypointNet categories: Ours versus KeypointDETR and prior detectors; baseline numbers as reported by KeypointDETR~\cite{jin2024keypointdetr}. Best per column in \textbf{bold}. $^{*}$All methods use the identical evaluation protocol, adopted from KeypointDETR.}
  \label{tab:keypointnet_percat_sota}
  \centering
  \begin{subtable}{\linewidth}
    \centering
    \caption{Hungarian mIoU$^{*}$ (\%) at $\epsilon = 0.10$.}
    \label{tab:keypointnet_percat_miou}
    \resizebox{\linewidth}{!}{%
    \begin{tabular}{@{}lrrrrrrrrrrrrrrrr@{}}
      \toprule
      Method & Air. & Bat. & Bed & Bot. & Cap & Car & Cha. & Gui. & Hel. & Kni. & Lap. & Mot. & Mug & Ska. & Tab. & Ves. \\
      \midrule
      PRA-Net~\cite{cheng2021net}             & 82.88 & 72.72 & 49.32 & 40.11 & 28.20 & 82.06 & 70.44 & 73.71 & 31.24 & 38.15 & 84.70 & 60.73 & 76.50 & 49.47 & 69.69 & 48.17 \\
      PointNet++~\cite{qi2017pointnet++}      & 82.31 & 71.77 & 57.85 & 50.77 & 45.78 & 84.06 & 57.50 & 72.34 & 48.51 & 62.57 & 87.20 & 69.93 & 68.75 & 67.67 & 60.01 & 58.52 \\
      DGCNN~\cite{wang2019dynamic}            & 78.49 & 60.33 & 64.30 & 56.79 & 47.56 & 79.12 & 59.49 & 63.88 & 46.79 & 70.19 & 85.22 & 68.69 & 63.23 & 76.78 & 59.92 & 54.56 \\
      Wei~\etal~\cite{wei2021multi}           & 79.86 & 65.50 & 52.05 & 55.58 & 36.14 & 86.13 & 86.28 & 78.08 & 47.72 & 63.66 & 94.40 & 69.38 & 70.79 & 72.91 & 86.61 & 58.38 \\
      KeypointDETR~\cite{jin2024keypointdetr} & 91.37 & 71.37 & 70.83 & 87.22 & 79.43 & \textbf{94.29} & 91.89 & 93.96 & \textbf{85.05} & 63.12 & 99.11 & 87.52 & \textbf{96.62} & 92.28 & \textbf{98.05} & 59.97 \\
      \midrule
      Ours                            & \textbf{96.93} & \textbf{90.83} & \textbf{87.61} & \textbf{90.68} & \textbf{91.74} & 92.25 & \textbf{94.58} & \textbf{97.84} & 52.36 & \textbf{91.12} & \textbf{99.65} & \textbf{92.28} & 85.55 & \textbf{94.25} & 95.96 & \textbf{77.35} \\
      \bottomrule
    \end{tabular}}
  \end{subtable}

  \begin{subtable}{\linewidth}
    \centering
    \caption{Chamfer Distance (CD$^{*}$, lower is better).}
    \label{tab:keypointnet_percat_cd}
    \resizebox{\linewidth}{!}{%
    \begin{tabular}{@{}lrrrrrrrrrrrrrrrr@{}}
      \toprule
      Method & Air. & Bat. & Bed & Bot. & Cap & Car & Cha. & Gui. & Hel. & Kni. & Lap. & Mot. & Mug & Ska. & Tab. & Ves. \\
      \midrule
      PRA-Net~\cite{cheng2021net}             & 0.049 & 0.082 & 0.182 & 0.163 & 0.238 & 0.059 & 0.060 & 0.053 & 0.170 & 0.188 & 0.064 & 0.099 & 0.070 & 0.163 & 0.053 & 0.146 \\
      PointNet++~\cite{qi2017pointnet++}      & 0.050 & 0.075 & 0.113 & 0.092 & 0.134 & 0.057 & 0.077 & 0.060 & 0.144 & 0.112 & 0.046 & 0.075 & 0.076 & 0.080 & 0.062 & 0.108 \\
      DGCNN~\cite{wang2019dynamic}            & 0.058 & 0.074 & 0.086 & 0.083 & 0.101 & 0.066 & 0.076 & 0.072 & 0.123 & 0.083 & 0.036 & 0.079 & 0.083 & 0.069 & 0.066 & 0.107 \\
      Wei~\etal~\cite{wei2021multi}           & 0.048 & 0.088 & 0.198 & 0.093 & 0.211 & 0.048 & 0.055 & 0.066 & 0.172 & 0.107 & 0.078 & 0.085 & 0.101 & 0.081 & 0.039 & 0.100 \\
      KeypointDETR~\cite{jin2024keypointdetr} & 0.045 & 0.074 & 0.081 & 0.050 & 0.066 & \textbf{0.045} & 0.055 & 0.045 & \textbf{0.087} & 0.089 & \textbf{0.024} & 0.053 & \textbf{0.049} & 0.055 & \textbf{0.027} & 0.105 \\
      \midrule
      Ours                            & \textbf{0.038} & \textbf{0.049} & \textbf{0.056} & \textbf{0.049} & \textbf{0.062} & 0.051 & \textbf{0.051} & \textbf{0.029} & 0.407 & \textbf{0.036} & 0.033 & \textbf{0.047} & 0.096 & \textbf{0.037} & 0.059 & \textbf{0.079} \\
      \bottomrule
    \end{tabular}}
  \end{subtable}
\end{table}

\subsection{Datasets}
\label{sec:datasets}

\paragraph{Building3D.} We use the corner-detection task of Building3D~\cite{wang2023building3d} on two splits, the Entry-Level subset and the full Tallinn city split. Predicted corners are matched with the Hungarian assignment~\cite{Kuhn1955Hungarian} to the ground truth and scored by Corner Precision, Recall, and F1 (CP, CR, CF1) and the Average Corner Offset (ACO, lower is better).

\paragraph{KeypointNet.} KeypointNet~\cite{you2020keypointnet} provides human-annotated semantic keypoints on ShapeNet objects across 16 categories. We follow the standard split and evaluate with the KeypointDETR~\cite{jin2024keypointdetr} evaluation protocol: Hungarian-matched mean intersection-over-union (mIoU) at a distance threshold $\epsilon$ and Chamfer Distance (CD). The mIoU and CD metrics are averaged over all test shapes, so the aggregate values in \cref{tab:postproc_ablation,tab:ablations} are shape-weighted rather than the mean of the per-category columns in \cref{tab:keypointnet_percat_sota}. The training regimes differ: KeypointDETR trains one model per category with a fixed keypoint set, and its numbers are taken from~\cite{jin2024keypointdetr}, whereas our model is trained on all 16 categories simultaneously. 

\begin{table}[tb]
  \caption{Our learned suppression versus heuristic post-processing (greedy NMS, DBSCAN) per backbone, on Building3D (Entry-Level, Tallinn) and KeypointNet (all 16 categories), all on the test splits. Best per backbone in \textbf{bold}.}
  \label{tab:postproc_ablation}
  \centering
  \resizebox{\linewidth}{!}{%
  \begin{tabular}{@{}ll rrrr rrrr rr@{}}
    \toprule
    & & \multicolumn{4}{c}{Building3D Entry-Level} & \multicolumn{4}{c}{Building3D Tallinn} & \multicolumn{2}{c}{KeypointNet} \\
    \cmidrule(lr){3-6}\cmidrule(lr){7-10}\cmidrule(lr){11-12}
    Backbone & Post-proc. & ACO$\downarrow$ & CP & CR & CF1 & ACO$\downarrow$ & CP & CR & CF1 & mIoU & CD$\downarrow$ \\
    \midrule
    Ours & NMS    & 0.285 & 92.3 & 89.2 & 90.70 & 0.310 & 85.4 & \textbf{84.8} & 85.10 & 84.25 & 0.080 \\
         & DBSCAN & 0.260 & 95.9 & 82.9 & 89.00 & 0.290 & \textbf{94.2} & 75.2 & 83.60 & 79.67 & 0.085 \\
         & Learned supp.  & \textbf{0.220} & \textbf{96.5} & \textbf{90.1} & \textbf{93.20} & \textbf{0.216} & 92.3 & 83.7 & \textbf{87.80} & \textbf{92.01} & \textbf{0.055} \\
    \midrule
    PointNet++~\cite{qi2017pointnet++} & NMS & 0.323 & 44.2 & \textbf{95.0} & 60.40 & 0.366 & 54.6 & \textbf{91.2} & 68.30 & 46.33 & 0.130 \\
         & DBSCAN & 0.277 & 89.5 & 76.1 & 82.30 & 0.312 & 90.2 & 72.5 & 80.40 & 70.92 & \textbf{0.111} \\
         & Learned supp.  & \textbf{0.270} & \textbf{89.9} & 84.2 & \textbf{87.00} & \textbf{0.295} & \textbf{92.4} & 78.5 & \textbf{84.90} & \textbf{74.99} & 0.115 \\
    \midrule
    DGCNN~\cite{wang2019dynamic} & NMS & 0.338 & 27.8 & \textbf{94.8} & 43.00 & 0.390 & 28.9 & \textbf{94.7} & 44.30 & 42.87 & 0.129 \\
         & DBSCAN & \textbf{0.287} & 82.1 & 75.0 & 78.40 & 0.337 & 77.5 & 72.4 & 74.90 & 70.68 & 0.100 \\
         & Learned supp.  & 0.295 & \textbf{90.7} & 79.8 & \textbf{84.90} & \textbf{0.305} & \textbf{89.7} & 75.0 & \textbf{81.70} & \textbf{77.87} & \textbf{0.099} \\
    \bottomrule
  \end{tabular}}
\end{table}

\subsection{Implementation Details}
\label{sec:impl}

\paragraph{Common setup.} Training follows the same two-stage procedure on both datasets. The feature propagation and graph layer MLPs use GELU, and the classification and offset heads use ReLU. The backbone is trained with AdamW under a OneCycleLR schedule with a peak learning rate of $3\times10^{-3}$; we then freeze it and train the learned suppression module with Adam. Candidate generation (\cref{sec:postproc}) uses vote threshold $0.10$ and NMS radius $0.05$; the module runs $T{=}2$ suppression rounds with radius $\rho{=}0.08$ and keep threshold $\tau{=}0.40$, and training matches candidates to keypoints by Hungarian assignment at threshold $0.10$. These settings are chosen on the validation split; the heuristic baselines in \cref{tab:postproc_ablation} use the same ones, greedy NMS being exactly this candidate generation stage and DBSCAN the configuration set in \cref{tab:ablations}a (eps $0.05$, threshold $0.19$), held fixed across backbones and datasets. The whole pipeline trains and runs on a single NVIDIA RTX~4070 (12\,GB VRAM), where the backbone holds roughly $22.7$M parameters and the suppression module adds $0.15$M. 

\paragraph{Building3D.} Each scene is subsampled to $2560$ points carrying XYZ, RGBA color, LiDAR intensity, and four scale-conditioning channels (the log normalization scale and bounding-box aspect ratios, broadcast to every point to recover the scale and shape lost to centroid-radius normalization), for $C{=}12$ input channels. The backbone is trained for $300$ epochs at batch size $32$ and the suppression module for $60$ epochs at batch size $8$, independently on the Entry-Level subset and on the full Tallinn city split. Training augments each scene with random flips and rotations, and soft labels (\cref{eq:sga}) use radius $r{=}0.05$. At inference we use test-time augmentation: each scene is passed through the network at four $90^{\circ}$ rotations, and a predicted corner is kept only if it is recovered in at least two of the four passes. 

\paragraph{KeypointNet.} A single backbone is trained jointly across all $16$ categories for $100$ epochs at batch size $14$, on shapes subsampled to $2048$ points carrying XYZ, color, and surface normals from principal component analysis (PCA) ($C{=}9$). Augmentation applies random flips, rotations, and scale jitter; soft labels use radius $r{=}0.10$. The suppression module uses the same settings as on Building3D.

\subsection{Experimental Results}
\label{sec:results}

\paragraph{Building3D.} On Building3D Entry-Level (\cref{tab:building3d_sota}) our detector reaches the best Corner Recall ($90.1$) and F1 ($93.2$) and matches the best ACO ($0.22$). It trails only PBWR, which fits parametric roof typologies, on Corner Precision ($96.5$ against $97.0$). On the larger Tallinn split our model remains competitive with BWFormer: it obtains the best Corner Recall ($83.7$ against $82.7$) and an F1 of $87.8$ against BWFormer's $88.4$.

\paragraph{KeypointNet.} On KeypointNet (\cref{tab:keypointnet_percat_sota}) a single model trained jointly on all 16 categories surpasses the per-category KeypointDETR on 12 of the 16 categories in mIoU, with the largest gains on categories such as Cap ($91.74$ against $79.43$), Bathtub ($90.83$ against $71.37$), and Vessel ($77.35$ against $59.97$). The Chamfer distances in \cref{tab:keypointnet_percat_cd} follow the same pattern. KeypointDETR remains ahead on a few categories, most notably Helmet, which we analyze below.

\paragraph{Effect of learned suppression.} \Cref{tab:postproc_ablation} isolates the contribution of the module by swapping only the post-processing stage, repeated on three backbones. On our backbone, learned suppression raises Corner F1 on Building3D Entry-Level from $90.70$ (NMS) and $89.00$ (DBSCAN) to $93.20$, and on Tallinn from $85.10$ and $83.60$ to $87.80$, while on KeypointNet it lifts mIoU from $84.25$ (NMS) and $79.67$ (DBSCAN) to $92.01$. The same pattern holds when the module is applied to a PointNet++ or a DGCNN backbone, where it gives the best F1 and mIoU in every case. This indicates that the improvement holds across both structural and semantic keypoints, using different backbones.

\paragraph{Component ablation.} The ablation in \cref{tab:ablations} builds the full model, starting from a PointNet++ base. On Building3D Entry-Level, the contrast GNN, the Point Transformer blocks, and the SA4 attention prior (which increases CP as spurious responses are suppressed, at the potential cost of a few weakly observed corners, lowering CR) together raise Corner F1 from $77.66$ to $84.32$, and the remaining refinements bring it to $88.55$. Replacing DBSCAN with the learned suppression module, first restricted to rescoring, adds $3.95$ F1 to reach $92.50$; enabling relocation leaves F1 mostly unchanged ($92.70$) but improves ACO from $0.280$ to $0.251$, showing that relocation contributes to localization accuracy. On KeypointNet (\cref{tab:ablations}b), moving from a Euclidean to a geodesic graph raises mIoU from $77.37$ to $79.67$; DBSCAN to greedy NMS lifts it to $84.25$, and the learned suppression module reaches $92.01$.

\paragraph{Failure analysis: Helmet.} Helmet is our weakest category: \cref{tab:keypointnet_percat_sota} reports $52.36$ mIoU and the largest Chamfer distance in the table ($0.407$), with prediction results in \cref{fig:qual_keypointnet}. This failure follows from the design of the pipeline: the detector consolidates candidates from local feature contrast and carries no category template, so it depends on landmarks that coincide with locally distinctive geometry. Helmet's landmarks are semantic conventions placed on a smooth shell with little such geometry. Cap, beside it in \cref{fig:qual_keypointnet}, has a similarly smooth global shape, but its landmarks sit on the brim, a sharp ridge, and the same trained model reaches $91.74$ mIoU there. KeypointDETR recovers Helmet ($85.05$ mIoU) with the per-category keypoint set that our single model forgoes.

\paragraph{Localization is bounded by candidate generation.} Our learned suppression module lowers the ACO over greedy NMS on every backbone (\cref{tab:postproc_ablation}). DBSCAN keeps two localization wins, ACO on DGCNN Entry-Level ($0.287$ against $0.295$) and CD on PointNet++ KeypointNet ($0.111$ against $0.115$); both come from candidate generation rather than suppression: DBSCAN averages each cluster while NMS keeps a single vote, so the cluster mean can begin closer to the true corner.
\section{Conclusion}
\label{sec:conclusion}

We presented a 3D keypoint detector built around a learned suppression module that improves over heuristic NMS and DBSCAN post-processing by suppressing and relocating coarse candidates over several rounds. The same architecture surpasses KeypointDETR on 12 of 16 KeypointNet categories, attains the best Corner Recall and F1 on Building3D Entry-Level, and remains competitive with BWFormer on Tallinn; the module improves Corner F1 and mIoU across all three backbones, including DBSCAN and non-maximum suppression. Two limitations remain: candidate generation still relies on a vote threshold and greedy NMS, and the KeypointDETR comparison spans different training regimes. Learning candidate generation, training the backbone and the suppression module jointly, and extending to scene-level clouds are natural next steps.

\bibliographystyle{splncs04}
\bibliography{main}
\end{document}